\documentclass[journal]{IEEEtran}

\usepackage{amsmath,graphicx}
\usepackage{subfig}
\usepackage{multirow}
\usepackage{color, soul}
\usepackage{diagbox}
\usepackage{multirow}
\usepackage{array}
\usepackage{fancyhdr}
\usepackage[
top=2cm,
bottom=2cm,
left=1.7cm,
right=1.7cm,
includefoot,
heightrounded, % to avoid spurious underfull messages
]{geometry} 
\newcolumntype{P}[1]{>{\centering\arraybackslash}p{#1}}
\newcolumntype{M}[1]{>{\centering\arraybackslash}m{#1}}

\begin{document}
\title{Scalable Learning with a Structural Recurrent 
	Neural Network for Short-Term Traffic Prediction}

\author{Youngjoo~Kim,
        Peng~Wang,
        and~Lyudmila~Mihaylova,~\IEEEmembership{Senior Member,~IEEE}% <-this % stops a space
\thanks{Y. Kim is with Sevendof AS, Trondheim, Norway. P. Wang and L. Mihaylova are with the Department of Automatic Control and Systems Engineering, University of Sheffield, Sheffield, United Kingdom (e-mail: rhymesg@gmail.com, peng.wang@sheffield.ac.uk, l.s.mihaylova@sheffield.ac.uk).}% <-this % stops a space
\thanks{Manuscript received ; revised .}}


\markboth{IEEE Sensors Journal, accepted manuscript to be published}%
{Kim \MakeLowercase{\textit{et al.}}: Manuscript for peer review}



\maketitle

\fancyhf{}% Clear all headers/footers
\renewcommand{\headrulewidth}{0pt}% No header rule
\renewcommand{\footrulewidth}{0pt}% No footer rule
\fancyfoot[L]{\textcopyright \:2019 IEEE. Personal use of this material is permitted. Permission from IEEE must be obtained for all other uses, in any current or future media, including reprinting/republishing this material for advertising or promotional purposes, creating new collective works, for resale or redistribution to servers or lists, or reuse of any copyrighted component of this work in other works.}%
\thispagestyle{fancy}%

\begin{abstract}
This paper presents a scalable deep learning approach for short-term traffic prediction based on historical traffic data in a vehicular road network. Capturing the spatio-temporal relationship of the big data often requires a significant amount of computational burden or an ad-hoc design aiming for a specific type of road network. To tackle the problem, we combine a road network graph with recurrent neural networks (RNNs) to construct a structural RNN (SRNN). The SRNN employs a spatio-temporal graph to infer the interaction between adjacent road segments as well as the temporal dynamics of the time series data. The model is scalable thanks to two key aspects. First, the proposed SRNN architecture is built by using the semantic similarity of the spatio-temporal dynamic interactions of all segments. Second, we design the architecture to deal with fixed-length tensors regardless of the graph topology. With the real traffic speed data measured in the city of Santander, we demonstrate the proposed SRNN outperforms the image-based approaches using the capsule network (CapsNet) by 14.1\% and the convolutional neural network (CNN) by 5.87\%, respectively, in terms of root mean squared error (RMSE). Moreover, we show that the proposed model is scalable. The SRNN model trained with data of a road network is able to predict traffic speed of different road networks, with the fixed number of parameters to train.
\end{abstract}


% \fi
\IEEEpeerreviewmaketitle

\section{Introduction}
\IEEEPARstart{M}{}aking accurate predictions of traffic data in a road network of interest is one of the important tasks for building intelligent transportation systems. The traffic data are usually obtained by magnetic induction loop detectors installed on road segments. The traffic data include traffic speed and flow, where the term traffic flow is used interchangeably with the terms traffic count or traffic volume. The sequence of traffic data on each road segment is essentially a time series, which is also spatially related to traffic in different road segments. Capturing the spatio-temporal features of the traffic data has been of great interest of researchers.

Deep learning approaches have recently been applied to traffic prediction tasks, given the advancement of technologies for obtaining and managing massive volume of traffic data. Early deep learning approaches to traffic prediction are based on convolutional neural networks (CNNs) \cite{Lv2015, Zhang2017}. They have been demonstrated to be effective in exploring spatial features. Studies incorporating recurrent neural networks (RNNs) have also been reported, considering the traffic prediction as a time series forecasting. Different gating mechanisms like long short-term memories (LSTMs) \cite{Zhang2017, Ma2015} and gated recurrent unit (GRU) \cite{Wu2018} have been employed in various architectures. A novel approach has been introduced in \cite{Ma2017} where the spatio-temporal traffic data are converted into images with two axes representing time and space and the images are fed into a CNN. This method enables the CNN model capture the spatio-temporal characteristics of the traffic data by learning the images. Recently, it has been demonstrated that a capsule network (CapsNet) architecture proposed in \cite{Kim2018} outperforms the CNN-based method in large, complex road networks. The dynamic routing algorithm of the CapsNet replaces the max pooling operation of the CNN, resulting in more accurate predictions but more parameters to train. Gaussian process (GP) is another data-driven approach, also known as a kernel-based learning algorithm. GPs have been demonstrated to be powerful in understanding the implicit relationship between data to give estimates for unseen points. Comparative studies \cite{Xie2010, Chen2015} have shown that GPs are effective in short-term traffic prediction. However, they still suffer from cubic time complexity in the size of training data \cite{Wang2018}.

The main contributions of this work are as follows: i) a structural RNN (SRNN) approach for traffic prediction that incorporates the topological information of the road network is proposed; ii) it is shown that the SRNN represents well both the spatial and temporal dynamics of the traffic; iii) the scalability of the proposed SRNN is demonstrated and validated with real data. Our work is mainly inspired by ideas from \cite{Jain2016, Vemula2018}. We build a spatio-temporal graph representation of the road network by considering each road segment as a node. All the nodes and edges are associated with RNNs that are jointly trained. It has been demonstrated in our preliminary work \cite{ICCASP} that the SRNN is more computationally efficient than the image-based state-of-the-art. In this paper, we introduce an advanced SRNN architecture with more details and provide a comprehensive demonstration for comparison with the image-based methods. More importantly, we show the proposed model is scalable; the SRNN model trained with a road network is capable of predicting traffic states in other road networks that have different network topologies (the number of road segments and how they are connected with one another). The proposed SRNN model is evaluated with the real dataset from the case studies of the SETA EU project \cite{SETA2016-2}. 

The rest of this paper is organized as follows. It starts with discussing the related work regarding deep learning methods for traffic prediction and SRNNs in Section II. We present the proposed architecture of the SRNN for traffic prediction in Section III. The proposed approach and results of the performance evaluation with a real dataset are given in Section IV. Finally, Section V discusses the results.

\section{Related Work}
In this section, we give a detailed overview of the relevant literature about previous works on deep learning for traffic prediction and SRNN. This will help readers figure out the background and differentiate our work from the existing works.

\subsection{Deep Learning for Traffic Prediction}

Recently, the notion of big data has been introduced to transportation research \cite{Zheng2016} and deep learning approaches have been actively used to address traffic prediction problems. One of the early approaches is based on a stacked auto-encoder model \cite{Lv2015}. Whereas this has a fully-connected structure, a method based on a CNN has been proposed \cite{Zhang2017}. However, this method treats the time dimension of the traffic flow as a channel of image data and therefore the temporal features of the traffic flow are ignored. The spatio-temporal features of the traffic have been of great interest of researchers.

One recent approach \cite{Ma2017} converts the traffic data into an image with two axes representing time and space and applies a CNN to capture the features of the spatio-temporal images. A CapsNet has also been used to improve the performance \cite{Kim2018} by effectively capturing the relationship between distant local features. A major drawback of these image-based methods is that the computational complexity increases as the number of road segments increases. Besides, a combination of a CNN, RNN, and attention model has been studied \cite{Wu2018} for mining spatial features, temporal features, and periodic features separately. However, this model aims for a simple network topology that consists of road segments along a straight path without intersections.

In this paper we develop a compact neural network model that is capable of learning spatio-temporal relationship regardless of the type of network topology. To broaden the applicability, we assume we are using only the historical traffic data and the topological information expressed as an adjacency matrix. The traffic data and the topological information are generally available for any road networks of interest.

\subsection{Structural Recurrent Neural Network}

The SRNN has been proposed in \cite{Jain2016} with the application in human motion forecasting, human activity anticipation, and driver maneuver anticipation where the connected components in a system of interest are represented as nodes in a graph. The SRNN is based on the spatio-temporal graphs which are usually used to model spatial and temporal reasoning \cite{Brendel2011}. The spatial and temporal interactions between nodes are parametrized with a factor graph \cite{Kschischang2001}. As the principles described in \cite{Jain2016} implies, the SRNN is applicable to any systems that can be expressed as spatio-temporal graphs.

An attention model has recently been applied to the SRNN \cite{Vemula2018} to find subsets of human crowds within which humans interact with each other, instead of using a fixed graph to represent the connection among the humans. This approach deals with only one kind of nodes, humans, to predict human trajectories, which resembles the problem of short-term traffic prediction.

Inspired by these two works \cite{Jain2016, Vemula2018} on the SRNN, we apply the SRNN to the short-term traffic prediction problem. 
We consider road segments as nodes that are semantically the same but use a road network topology to construct the spatio-temporal graph. Moreover, we verify the scalability of the SRNN by using training and evaluation datasets that have different network topologies.

\section{Structural Recurrent Neural Network for Scalable Traffic Prediction} \label{sec:SRNN}

\subsection{Problem Definition}
In this study, we address the problem of short-term traffic speed prediction based on historical traffic speed data and a road network graph. Suppose we deal with $N$ road segments where the loop detectors are installed. Let $x_u^t$ represent the traffic speed on road segment $u$ at time step $t$. Given a sequence of traffic speed data $\{x_u^t\}$ for road segments $u = 1, 2, ..., N$ at time steps $t = t_c-l+1, ..., t_c$, we predict the future traffic speed $x_u^{t_c+1}$ on each road segment where $t_c$ denotes the current time step and $l$ denotes the length of a historical data sequence under consideration. The road network graph is denoted as $\mathcal{G}=(\mathcal{V}, \mathcal{E})$ where $\mathcal{G}$ denotes the graph, $\mathcal{V}$ denotes the set of nodes, and $\mathcal{E}$ denotes the set of edges connecting two nodes in $\mathcal{V}$. In this study, the nodes in the graph correspond to road segments of interest. Thus, $|\mathcal{V}|=N$. We use a directed adjacency matrix $\mathcal{A}$ to represent the edges in the graph. For example, suppose the traffic flow comes from node $u$ to node $v$. This means there is an edge $e$ linking nodes $u$ and $v$. Thus, $\mathcal{A}(u,v) = 1$ and $e = (u,v) \in \mathcal{E}$. We say two nodes $u$ and $v$ are connected if either $(u,v) \in \mathcal{E}$ or $(v,u) \in \mathcal{E}$. Note that we use $e$ to denote an edge in general or the form of $(u,v)$ with parentheses to reveal the nodes connected by the edge.

\subsection{Spatio-Temporal Graph Representation} \label{sec:stgraph}

\begin{figure}[t!]
	\centering
	\subfloat[Spatio-temporal graph]{%
		\includegraphics[clip,width=0.9\linewidth]{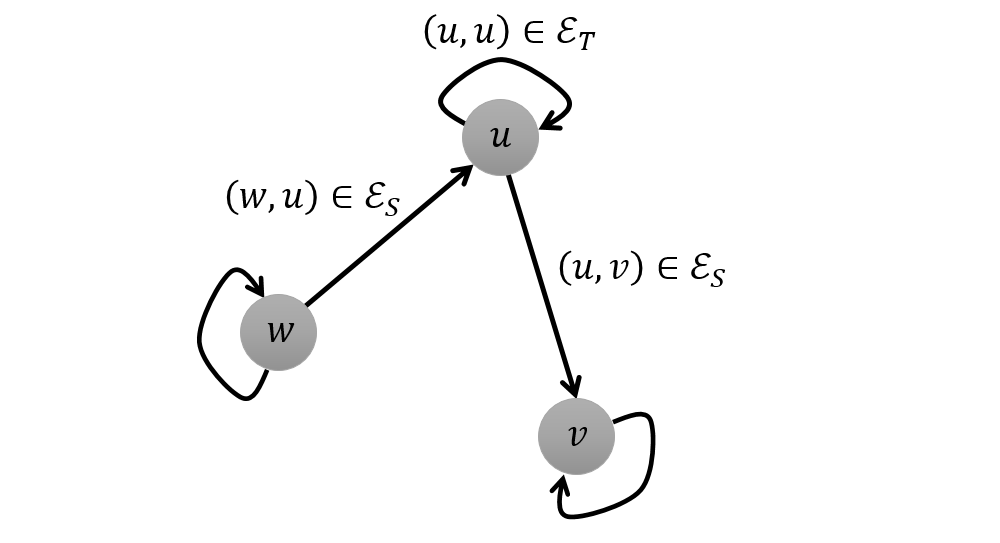}%
	}
	
	\subfloat[Unrolled over time]{%
		\includegraphics[clip,width=0.9\linewidth]{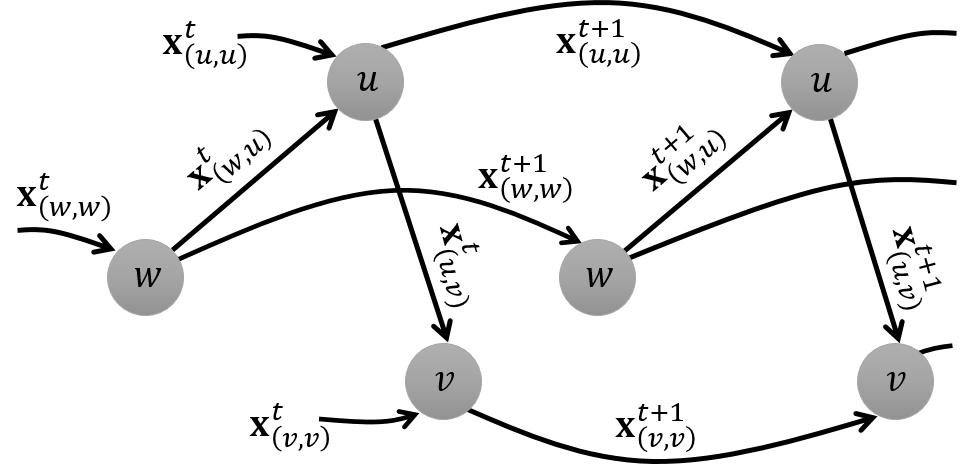}%
	}
	
	\caption{An example spatio-temporal graph. (a) Nodes represent road segments and the nodes are linked by edges in $\mathcal{E}_S$ and temporal edges in $\mathcal{E}_T$. (b) The spatio-temporal graph that is unrolled over time through the temporal edges in $\mathcal{E}_T$. The edges are labelled with the corresponding feature vectors.}
	\label{fig:stgraph}
\end{figure}

The SRNN is constructed based on a spatio-temporal graph that is obtained by unrolling the spatial graph $\mathcal{G}$ over time. We use a spatio-temporal graph representation $\mathcal{G}_{ST}=(\mathcal{V}, \mathcal{E}_S, \mathcal{E}_T)$. Let $\mathcal{G}_{ST}$ denote the spatio-temporal graph. $\mathcal{V}$, $\mathcal{E}_S$, and $\mathcal{E}_T$ denote the set of nodes, the set of spatial edges, and the set of temporal edges, respectively. Note that $\mathcal{E} = \mathcal{E}_S$.

The spatial edges in $\mathcal{E}_S$ represent the dynamics of traffic interaction between two adjacent road segments, and the temporal edges in $\mathcal{E}_T$ represent the dynamics of the temporal evolution of the traffic speed in road segments. Fig. \ref{fig:stgraph}(a) shows an example spatio-temporal graph. Nodes $u, v, w \in \mathcal{V}$ represent road segments. The connections between the road segments are represented by spatial edges in $\mathcal{E}_S$. Note that our approach differs from \cite{Vemula2018} in that the spatial edges are established if the two road segments are connected, whereas \cite{Vemula2018} employs an attention model on a fully-connected graph. Besides, a temporal edge originated from node $u$ is pointing to node $u$. The spatial graph $(\mathcal{V}, \mathcal{E}_S)$ is unrolled over time using temporal edges in $\mathcal{E}_T$ to form $\mathcal{G_{ST}}$ as depicted in Fig. \ref{fig:stgraph}(b) where the edges are labelled with the corresponding feature vectors.

\begin{figure}[t!]
	\centering
	\includegraphics[clip,width=0.95\linewidth]{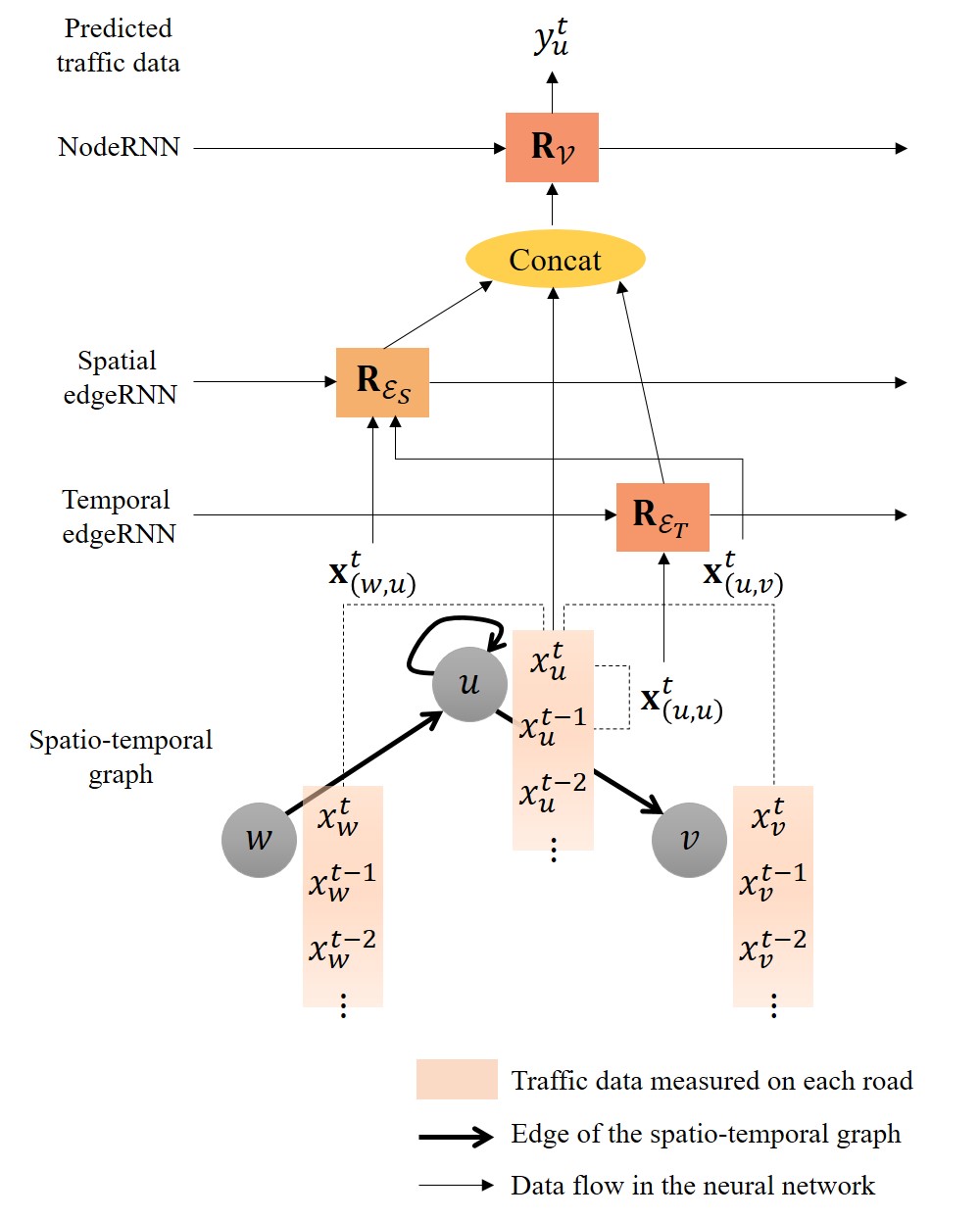}%
	\caption{Architecture of the SRNN in perspective of node $u$ drawn with the spatio-temporal graph.}
	\label{fig:SRNN}
\end{figure}

The feature of node $u \in \mathcal{V}$ at time step $t$ is $x_u^t$, denoting the traffic speed on the road segment. The feature vector of spatial edge $(u,v) \in \mathcal{E}_S$ at time step $t$ is $\mathbf{x}_{(u,v)}^t = [x_u^t, x_v^t] $, which is obtained by concatenating the features of nodes $u$ and $v$. As explained above, the spatial edge $(u,v)$ means that the traffic flow comes from node $u$ to node $v$. If the traffic flow goes in the opposite direction, the edge will be denoted as $(v,u)$ and associated with the different feature vector, $\mathbf{x}_{(v,u)}^t = [x_v^t, x_u^t]$. Note that we employ one spatial edge for each link between two nodes whose feature vector suffices to represent the directionality of the interaction between the nodes, as opposite to our preliminary work \cite{ICCASP} using two spatial edges for each link. The feature vector of temporal edge $(u,u) \in \mathcal{E}_T$ at time step $t$ is $\mathbf{x}_{(u,u)}^t = [x_u^{t-1}, x_u^t] $, which is obtained by concatenating the features of node $v$ at the previous time step and the current time step.

\subsection{Model Architecture}

In our SRNN architecture, the sets of nodes $\mathcal{V}$, spatial edges $\mathcal{E}_S$, and temporal edges $\mathcal{E}_T$ are associated with RNNs denoted as nodeRNN, $\mathbf{R}_{\mathcal{V}}$, spatial edgeRNN, $\mathbf{R}_{\mathcal{E}_S}$, and temporal edgeRNN, $\mathbf{R}_{\mathcal{E}_T}$, respectively. The SRNN is derived from the factor graph representation \cite{Jain2016}. Our architecture is the simplest case where the nodes, spatial edges, and temporal edges are sharing the same factors, respectively. This means we assume the dynamics of the spatio-temporal interactions is semantically same for all road segments, which keeps the overall parametrization compact and makes the architecture scalable and independent of the road network topology. Readers interested in the factor graph representation can refer to \cite{Kschischang2001}.

Fig. 2 visualizes the overall architecture. For each node $u$, a sequence of node features $\{x_u^t\}_{t=t_c-l+1}^{t_c}$ is fed into the architecture. Every time each node feature enters, the SRNN is supposed to predict the node label $y_u^{t}$, which corresponds to the traffic speed at the next time step, $x_u^{t+1}$. The input into the edgeRNNs is the edge feature $\mathbf{x}_{e}^t$ of edge $e \in \mathcal{E}_S \cup \mathcal{E}_T$, where the edge is incident to node $u$ in the spatio-temporal graph. The node feature $x_u^t$ is concatenated with the outputs of the edgeRNNs to be fed into the nodeRNN. We use LSTMs for the RNNs.

\subsection{Forward Path Algorithm} \label{sec:forward}

In this subsection, we provide a detailed explanation on how our SRNN algorithm works in a forward path. Before we run the forward path, we assure the adjacency matrix, $\mathcal{A}$, describing the road network topology is read to form the set of spatial edges, $\mathcal{E}_S$. The set of spatial edges connected to node $u$ is then obtained as $\mathcal{C}(u) = \{(u,v) \in \mathcal{E}_S \: or \: (v,u) \in \mathcal{E}_S,\: \forall v \in \mathcal{V}\}$.

For each time step $t$, the node feature, $x_u^t$ for $u \in \mathcal{V}$, the spatial edge feature, $\mathbf{x}_{e}^t$ for $e \in \mathcal{E}_S$, the temporal edge feature, $\mathbf{x}_{e}^t$ for $e \in \mathcal{E}_T$, are fed into the architecture. The inputs to the RNNs are converted into fixed-length vectors by an embedding function, denoted as $\phi(\cdot)$. The embedding function applies a linear transformation to the input with a rectified linear unit (ReLU) activation and dropout. For each node $u$, the following RNNs are executed to predict the node label $y_u^t$.

\subsubsection{Spatial EdgeRNN}

The spatial edgeRNN, $\mathbf{R}_{\mathcal{E}_S}$, takes the spatial edge features, $\{\mathbf{x}_{e}^t\}_{e \in \mathcal{E}_S}$, and the hidden state, $h_{S}^{t-1}$, as inputs. The hidden state, $h_S^{t-1}$, has a dimension of $|\mathcal{E}_S| \times \lambda_{\mathcal{E}_S}$ where $\lambda_{\mathcal{E}_S}$ is the size of the spatial edgeRNN. The spatial edge features are fed into a LSTM cell after converted into a fixed-length vector $\mathbf{a}_{S}^t$ by an embedding function as:
\begin{equation}
\mathbf{a}_{S}^t = \phi(\{ \mathbf{x}_{e}^t \}_{e \in \mathcal{E}_S} ; W_S^E)
\end{equation}
\begin{equation}
h_S^t = \textrm{LSTM}(\mathbf{a}_{S}^t, h_S^{t-1}; W_S^L)
\end{equation}
where $\phi(\cdot)$ denotes the embedding function, $W_S^E$ denotes the weight associated with the embedding function, and $W_S^L$ denotes the weight associated with the LSTM cell.

\subsubsection{Temporal EdgeRNN}
The temporal edgeRNN, $\mathbf{R}_{\mathcal{E}_T}$, takes the temporal edge features, $\{\mathbf{x}_{e}^t\}_{e_T \in \mathcal{E}_T}$, and the hidden state, $h_{T}^{t-1}$, as inputs. The hidden state, $h_T^{t-1}$, has a dimension of $|\mathcal{E}_T| \times \lambda_{\mathcal{E}_T}$ where $\lambda_{\mathcal{E}_T}$ is the size of the temporal edgeRNN. Similarly to the spatial edgeRNN, the temporal edge features go through the embedding function and a LSTM cell as:
\begin{equation}
\mathbf{a}_{T}^t = \phi(\{ \mathbf{x}_{e}^t \}_{e \in \mathcal{E}_T} ; W_T^E)
\end{equation}
\begin{equation}
h_T^t = \textrm{LSTM}(\mathbf{a}_{T}^t, h_T^{t-1}; W_T^L)
\end{equation}
where $\phi(\cdot)$ denotes the embedding function, $W_T^E$ denotes the weight associated with the embedding function, and $W_T^L$ denotes the weight associated with the LSTM cell.

\subsubsection{NodeRNN}

The nodeRNN, $\mathbf{R}_{\mathcal{V}}$, takes the node features, $\{x_u^t\}_{u \in \mathcal{V}}$, the hidden state, $h^{t-1}$, and the outputs of the spatial edgeRNN and the temporal edgeRNN as inputs. 

For each node $u \in \mathcal{V}$, rows of the hidden state of the spatial edgeRNN associated with the spatial edges connected to node $u$ are selected as:
\begin{equation}
h_{\mathcal{C}(u)}^t = h_S^t(\mathcal{C}(u))
\end{equation}
where $h_{\mathcal{C}(u)}^t$ has a dimension of $|\mathcal{C}(u)| \times \lambda_{\mathcal{E}_S}$. We add up the row vectors of $h_{\mathcal{C}(u)}^t$ to get the spatial edges' contribution to the nodeRNN as:
\begin{equation}
h_{S_u}^t = \mathrm{sum}(h_{\mathcal{C}(u)}^t)
\end{equation}
where $h_{S_u}^t$, encapsulating the average spatial influence to node $u$, has a dimension of $1 \times \lambda_{\mathcal{E}_S}$. We employ summation, rather than concatenation, here to realize a fixed architecture regardless of the graph topology. This is one of key efforts to control the number of parameters by representing the variable context with a fixed-length vector.
A row vector of the hidden state of the temporal edgeRNN associated with node $u$, denoted as $h_{T_u}^t$, is concatenated with $h_{S_u}^t$ to form
\begin{equation}
H_{u}^t = \mathrm{concat}(h_{T_u}^t, h_{S_u}^t)
\end{equation}
where $H_{u}^t$ has a dimension of $1 \times (\lambda_{\mathcal{E}_T}+\lambda_{\mathcal{E}_S})$. Vertically concatenating these row vectors for all nodes in $\mathcal{V}$ provides an input to the nodeRNN, denoted as $H^t$. Each row of $H^t$ represents the influence of the spatio-temporal edges to the corresponding node feature. 

The node features, $\{x_u^t\}_{u \in \mathcal{V}}$, and the concatenated hidden state, $H^t$, are fed into a LSTM cell after converted into fixed-length vectors by embedding functions as:
\begin{equation}
\mathbf{a}^t = \phi(\{x_u^t\}_{u \in \mathcal{V}}; W^E)
\end{equation}
\begin{equation}
\mathbf{a}_H^t = \phi(H^t; W_H^E)
\end{equation}
\begin{equation}
h^t = \textrm{LSTM}(\mathrm{concat}(\mathbf{a}^t, \mathbf{a}_H^t), h^{t-1}; W^L)
\end{equation}
\begin{equation}
\{y_u^t\} = W^O h^t
\end{equation}
where $\phi(\cdot)$ denotes the embedding function, $W^E$ denotes the weight associated with the embedding function for node features, $W_H^E$ denotes the weight associated with the embedding function for the concatenated hidden state, and $W^L$ denotes the weight associated with the LSTM cell. The output hidden state of the nodeRNN is passed through a linear layer with $W^O$ to provide the output node features $\{y_u^t\}$ that represent the predicted labels for all nodes.

\subsection{Training the Structural Recurrent Neural Network} \label{sec:tranining_network}

In order to train the SRNN model, the traffic speed measurements for every time step $t$ are used to comprise the edge features, $\{ \mathbf{x}_{e}^t \}_{e \in \mathcal{E}_S}$ and $\{ \mathbf{x}_{e}^t \}_{e \in \mathcal{E}_T}$, and the node features $\{x_u^t\}_{u \in \mathcal{V}}$, as described in Section \ref{sec:stgraph}. The directional adjacency matrix, $\mathcal{A}$, is read to form $\mathcal{C}(u)$ for $u \in \mathcal{V}$.

After the forward path addressed in Section \ref{sec:forward} is executed, the prediction error, $\{x_u^{t+1}\}-\{y_u^t\}$, is jointly back-propagated through the nodeRNN, the spatial edgeRNN, and the temporal edgeRNN involved in the forward path. We employ mean squared error (MSE) as a loss function.

Note that the trainable parameters of this SRNN are $\{W_S^E, W_S^L, W_T^E, W_T^L, W^E, W_H^E, W^L, W^O\}$ whose size is independent of the size of the spatio-temporal graph. It only depends on the sizes of the RNNs.

\section{Validation with Real Data}

This section presents the proposed approach and results of performance validation. The purpose of the experiment is to demonstrate: 1) the SRNN outperforms the image-based state-of-the-art methods, CapsNet \cite{Kim2018} and CNN \cite{Ma2017}, in learning the spatially-related time series data; and 2) the SRNN is scalable regardless of the network topology. The detailed methods for preparing the dataset and the neural networks are described in Section \ref{sec:implementation}.

\subsection{Implementation Details}\label{sec:implementation}

\begin{figure}[t!]
	\centering
	\includegraphics[clip,width=0.95\linewidth]{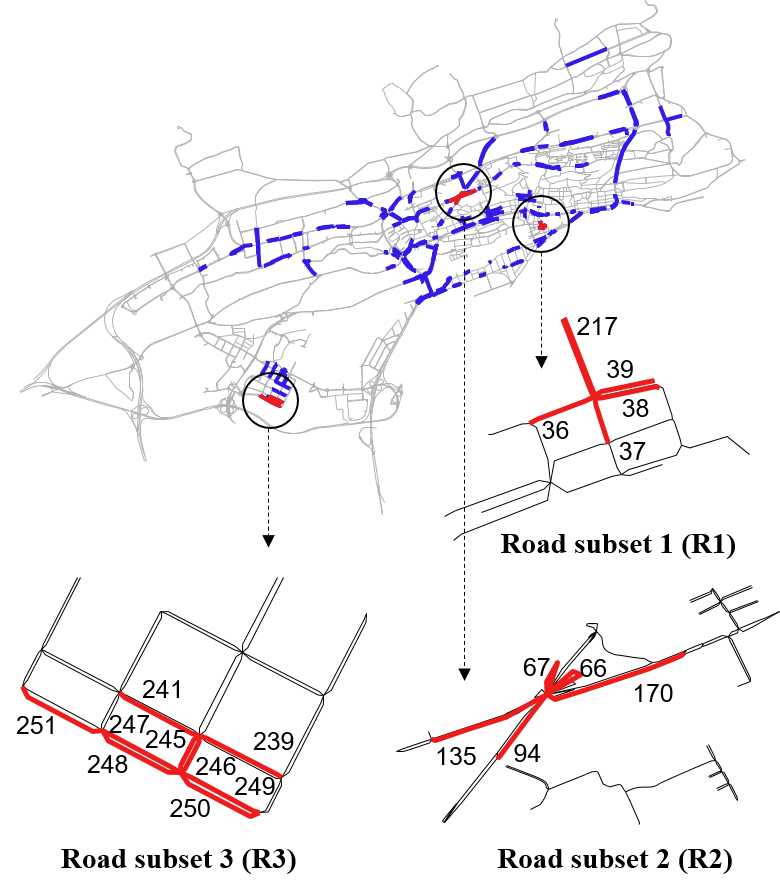}%
	
	\caption{Road segments marked in blue and red in the road network are the road segments of interest in which the sensors are installed and the traffic data are available. Road segments marked in red are belong to road subsets used in the performance validation. The road networks around the three road subsets are magnified below. Next to each road segment is a unique ID number for the segment.}
	\label{fig:road}
\end{figure}

In this subsection, we present how datasets are set up and how the different neural networks are implemented for the performance validation.

\subsubsection{Datasets} \label{sec:datasets}

We use the traffic speed data that had been gathered in the city of Santander, Spain during the year of 2016, which are available from the case studies of the SETA EU project \cite{SETA2016-2}. Traffic speed measurements taken by magnetic loop detectors on road segments are aggregated every 15 minutes and each sparsely missing measurement is masked with an average of the speed data recorded at the same time in the other days. The overall dataset contains 35,040 speed readings per road segment. We use data of the first 9 months as the training set and the remaining data of the last 3 months as the evaluation set. We scale the traffic speed data into the range [0,1] before feeding into the neural networks. In addition, the SRNN uses the adjacency matrix that represents the directional connections between the road network. The adjacency matrix for the road segments with sensors installed, which are of our interest, is extracted from the adjacency matrix for the whole network by taking only the rows and columns associated with the road segments with sensors.

On top of Fig. \ref{fig:road} is the road network of the city of Santander. The blue lines denote the road segments of interest where the traffic data are available. Since such road segments are sparsely located, the road network of interest is a disconnected graph where there is at least one unreachable node, or road segment, starting from a different node in the graph. The road network consists of many small connected graphs whose size varies from 1 to 9. We deal with these connected road subsets because we believe traffic data from the adjacent road segments are correlated, which is referred to as their spatial relationship to be learnt by the SRNN. We use three largest connected sets for validation and denote them as Road subset 1 (R1), Road subset 2 (R2), and Road subset 3 (R3), as shown in Fig. \ref{fig:road}. R1 and R2 have 5 road segments and R3 consists of 9 road segments. Next to each road segment is a unique ID number for the segment. 

\begin{figure}[t!]
	\centering
	\subfloat[Road subset 1 (R1)]{%
		\includegraphics[clip,width=0.475\linewidth]{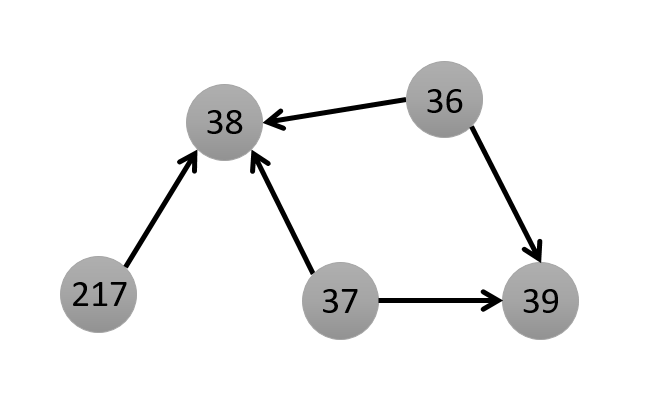}%
	}
	\hfill
	\subfloat[Road subset 2 (R2)]{%
		\includegraphics[clip,width=0.475\linewidth]{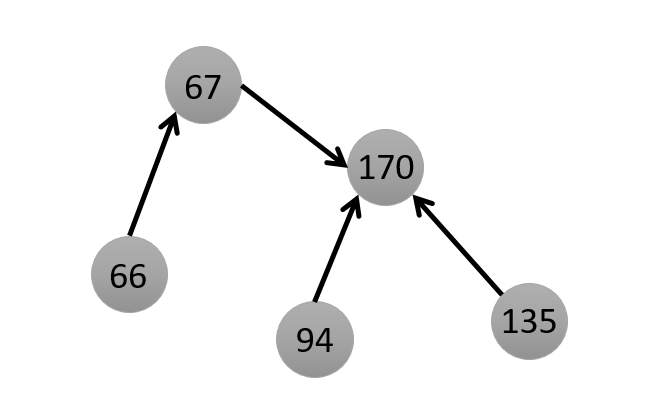}%
	}
	\vskip\baselineskip
	\centering
	\subfloat[Road subset 3 (R3)]{%
		\includegraphics[clip,width=0.475\linewidth]{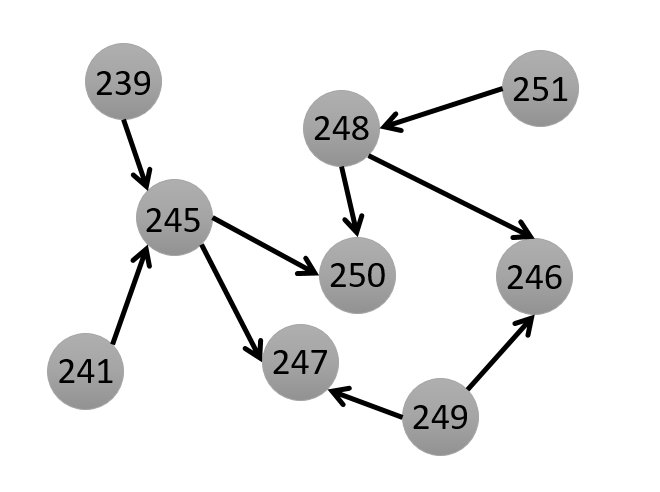}%
	}
	\hfill
	\subfloat[Road subset 4 (R4)]{%
		\includegraphics[clip,width=0.475\linewidth]{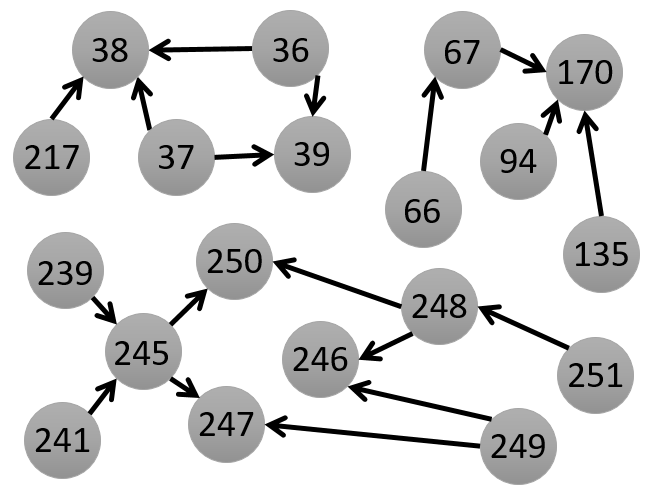}%
	}
	\caption{Graph representation of four road subsets used in the performance validation. Nodes in the shape of grey circle, with the corresponding ID numbers, denotes road segments. The arrows denote spatial edges that represent directional connection between two adjacent road segments. R4 is the sum of sets R1, R2, and R3. ID numbers for the road segments are presented in the nodes.}
	\label{fig:road_subset}
\end{figure}

In the experiment, we use traffic speed data and adjacency matrices for 4 road subsets including Road subset 4 (R4) that is the sum of sets R1, R2, and R3. Their graph representation illustrated in Fig. \ref{fig:road_subset} clarifies the spatial connections between the nodes.

\subsubsection{Neural Networks}

We compare the performance of the traffic speed prediction provided by the 3 different neural networks:
\begin{itemize}
\item Proposed SRNN
\item CapsNet \cite{Kim2018}
\item CNN \cite{Ma2017}
\end{itemize}
Here we provide a detailed description of how these neural networks are implemented. All the neural networks employ MSE as a loss function and Adam optimizer \cite{Kingma2014} to minimize the sum of the MSE. 

The architecture of the proposed SRNN can be referred to Section \ref{sec:SRNN}. The SRNN shows its best performance with the following hyperparameters. The hidden state of the nodeRNN has a size of 64, and that of the edgeRNNs has a size of 64 as well. We employ embedding layers in the network that convert the inputs into a 32-dimensional vectors with a dropout rate of 0.5.  SRNN is trained with the starting learning rate of 0.0005 and the exponential decay rate of 0.99. The major hyperparameters of the SRNN are summarized in Table \ref{table:SRNN}. The SRNN is built based on the Pytorch implementation of \cite{Vemula2018}. The computational complexity of the SRNN is $\mathcal{O}(W)$ since the SRNN consists of 3 RNNs (a nodeRNN, a spatial edgeRNN, and a temporal edgeRNN executed sequentially) and the computational complexity of a RNN with a LSTM is $\mathcal{O}(W)$ \cite{Sak2014} where $W$ is the number of trainable parameters. Note that the number of trainable parameters is independent of the size of the road network as discussed in Section \ref{sec:tranining_network}.

The CapsNet and CNN are built based on the Tensorflow implementation of \cite{Kim2018}. In summary, they convert the spatio-temporal traffic data into an image with two axes representing space and time and apply the deep learning methods for capturing the relationship between the spatio-temporal features in the image. Only one max pooling operation is used in the CNN implementation because the spatio-temporal images from the datasets are small (e.g., 5 segments for R1 and R2). Both of the networks show their best performance with the common starting learning rate of 0.0005 and the exponential decay rate of 0.9999. The computational complexity of the CNN is proportional to the spatial size of the output feature map \cite{He2015}, which is proportional to the number of road segments in our problem, given the number of filters and the filter size are constant. Thus, the CNN has $\mathcal{O}(N)$. On the other hand, the computational complexity of the CapsNet will be proportional to $\mathcal{O}(N^2)$ since it runs routing-by-agreement between capsule features in two convolutional layers \cite{Sabour2017}.

\begin{table}[t!]
	\captionsetup{justification=centering}
	\caption{Hyperparameters of the proposed SRNN.}
	\centering
	\renewcommand{\arraystretch}{1.5}
	\begin{tabular}{|M{3.5cm}|M{1.0cm}|}
		\hline
		\textbf{Hyperparameter} & \textbf{Value} \\ \hline
		NodeRNN size, $\lambda$ & 64 \\ \hline
		Spatial edgeRNN size, $\lambda_{\mathcal{E}_S}$ & 64   \\ \hline
		Temporal edgeRNN size, $\lambda_{\mathcal{E}_T}$ & 64     \\ \hline
		Embedding size  & 32    \\ \hline
		Learning rate & 0.0005   \\ \hline
		Decay rate & 0.99     \\ \hline
		Dropout rate  & 0.5    \\ \hline
		
	\end{tabular}
	\renewcommand{\arraystretch}{1}
	\label{table:SRNN}
\end{table}

\subsubsection{Task}

Given the datasets, the neural networks are supposed to give 15-min prediction of traffic speed based on 150-min data, which corresponds to a short-term prediction for the next time step based on data of previous 10 time steps $(l=10)$.

\subsubsection{Performance Metrics}
We use the root mean squared error (RMSE) as a performance metric for assessing the prediction accuracy as:
\begin{equation}
RMSE = \sqrt { \frac{\sum_{i=1}^{I}{ (y^{(i)} - \hat{y}^{(i)})^2 }}  {I} }
\end{equation}
where $\hat{y}^{(i)}$ and $y^{(i)}$ denote the $i$-th speed prediction and its true value, respectively. Here, $I$ represents the number of the speed data in the evaluation set. Although there are other metrics such as mean relative error (MRE) and mean absolute error (MAE), we believe the RMSE suffices to show the performance difference between the neural networks employing MSE as a loss function. The MRE and MAE show a similar tendency as the RMSE in this experiment \cite{ICCASP}.

Besides the accuracy, we compare the number of trainable parameters of the neural networks. Since computation time depends on the type of software platform, computing machine, code optimization, and the usage of graphics processing unit (GPU), we see the number of trainable parameters as an empirical measure of the computational complexity. The number of trainable parameters is obtained by counting the number of variables that are optimized by back-propagation, such as the weights and biases in the neural networks.

\subsection{Learning Spatio-Temporal Relationship} \label{sec:learning_st}

\begin{table}[t!]
	\captionsetup{justification=centering}
	\caption{Prediction performance in RMSE (unit: km/h).}
	\centering
	\renewcommand{\arraystretch}{1.5}
	\begin{tabular}{|M{1.0cm}|M{0.7cm}|M{0.7cm}|M{0.7cm}|M{0.7cm}|}
		\hline
		& \textbf{R1} & \textbf{R2} & \textbf{R3} & \textbf{R4} \\ \hline
		\textbf{SRNN} & 6.877         & 8.537         & 9.419        & 9.322         \\ \hline
		\textbf{CapsNet}  & 9.416         & 9.931         & 10.52         & 9.910         \\ \hline
		\textbf{CNN}  & 7.329         & 9.075         & 10.54         & 10.36         \\ \hline
		
	\end{tabular}
	\renewcommand{\arraystretch}{1}
	\label{table:learning_st}
\end{table}

Here we verify the SRNN is capable of learning the spatio-temporal relationship accurately, requiring fewer parameters to train. Whereas the RNN learns the time series data on each road as if data on different roads are independent, the SRNN takes the road graph as well as the historical traffic data and learns the spatial relationship between data from different road segments and temporal relationship of time series data on each road.

\begin{table}[t!] 
	%\captionsetup{justification=centering}
	\caption{Prediction performance in RMSE, trained and evaluated with different road networks (unit: km/h).}
	\centering
	\renewcommand{\arraystretch}{2}
	\subfloat[SRNN]{%
		\begin{tabular}{|M{2.2cm}|M{0.7cm}|M{0.7cm}|M{0.7cm}|M{0.7cm}|}
			\hline
			\backslashbox{\textbf{Train}}{\textbf{Eval}}
			& \textbf{R1}   & \textbf{R2}   & \textbf{R3}   & \textbf{R4}   \\ \hline
			\textbf{R1} & 6.877 & 8.697 & 9.504 & 9.455 \\ \hline
			\textbf{R2} & 6.990 & 8.537 & 9.653 & 9.444 \\ \hline
			\textbf{R3} & 7.141 & 9.450 & 9.419 & 10.06 \\ \hline
			\textbf{R4} & 6.871 & 8.714 & 9.400 & 9.322 \\ \hline
		\end{tabular}
	}
	
	\subfloat[CapsNet]{%
		\begin{tabular}{|M{2.2cm}|M{0.7cm}|M{0.7cm}|M{0.7cm}|M{0.7cm}|}
			\hline
			\backslashbox{\textbf{Train}}{\textbf{Eval}}
			& \textbf{R1}   & \textbf{R2}   & \textbf{R3}   & \textbf{R4}   \\ \hline
			\textbf{R1} & 9.416 & 23.46 & - & - \\ \hline
			\textbf{R2} & 21.26 & 9.931 & - & - \\ \hline
			\textbf{R3} & - & - & 10.52 & - \\ \hline
			\textbf{R4} & - & - & - & 9.910 \\ \hline
		\end{tabular}
	}
	
	\subfloat[CNN]{%
		\begin{tabular}{|M{2.2cm}|M{0.7cm}|M{0.7cm}|M{0.7cm}|M{0.7cm}|}
			\hline
			\backslashbox{\textbf{Train}}{\textbf{Eval}}
			& \textbf{R1}   & \textbf{R2}   & \textbf{R3}   & \textbf{R4}   \\ \hline
			\textbf{R1} & 7.329 & 23.57 & - & - \\ \hline
			\textbf{R2} & 18.77 & 9.075 & - & - \\ \hline
			\textbf{R3} & - & - & 10.54 & - \\ \hline
			\textbf{R4} & - & - & - & 10.36 \\ \hline
		\end{tabular}
	}
	
	\label{table:scalability}
	\renewcommand{\arraystretch}{1}
\end{table}

The three neural networks are trained and evaluated with the datasets R1, R2, R3, and R4. Each dataset contains the 1-year traffic speed data, where the data of the first 9 months constitute the training set and the remaining data of the last 3 months are taken as the evaluation set. We run 10 epochs of training and evaluation and take the average performance as the result. Table \ref{table:learning_st} shows the average prediction performance of the methods on each dataset. One can observe that the CNN performs better than the CapsNet for datasets R1 and R2, but worse for datasets R3 and R4. This observation confirms the result reported in \cite{Kim2018} that the CapsNet is more effective when the image size (the number of road segments) is larger. Meanwhile, the SRNN outperforms both of the image-based methods. In average, the SRNN is better than the CapsNet and the CNN by $14.1 \%$ and $5.87 \%$, respectively, in RMSE. Note that the performance is realized with the smaller number of parameters to train. The number of trainable parameter of the SRNN is $8.79 \times 10^4$ whereas that of the CapsNet ranges from $5.59 \times 10^5$ (R1 and R2) to $7.44 \times 10^6$ (R4) and that of the CNN ranges from $3.75 \times 10^5$ (R1 and R2) to $4.26 \times 10^5$ (R4). 

\subsection{Scalability to Network Topology} \label{sec:scalability}

In order to demonstrate the scalability, we evaluate the three neural networks by using datasets different from the dataset used in training. For example, the neural networks trained with the training set (data of the first 9 months) of R1 are evaluated with the evaluation set (data of the last 3 months) of the other datasets: R2, R3, and R4. This is repeated for evaluations with training sets of R2, R3, and R4, respectively.

Table \ref{table:scalability} summarizes the result. The number in $i$-th row and $j$-th column denotes the RMSE obtained with the training set of R$i$ and the evaluation set of R$j$. The diagonal terms are equal to the result in Table \ref{table:learning_st}. The SRNN shows consistent performance for all the combinations of training sets and evaluation sets. The evaluation performance tends to depend on the dataset used in evaluation, regardless of the dataset used in training.

\begin{figure}[t!]
	\centering
	\includegraphics[clip,width=0.95\linewidth]{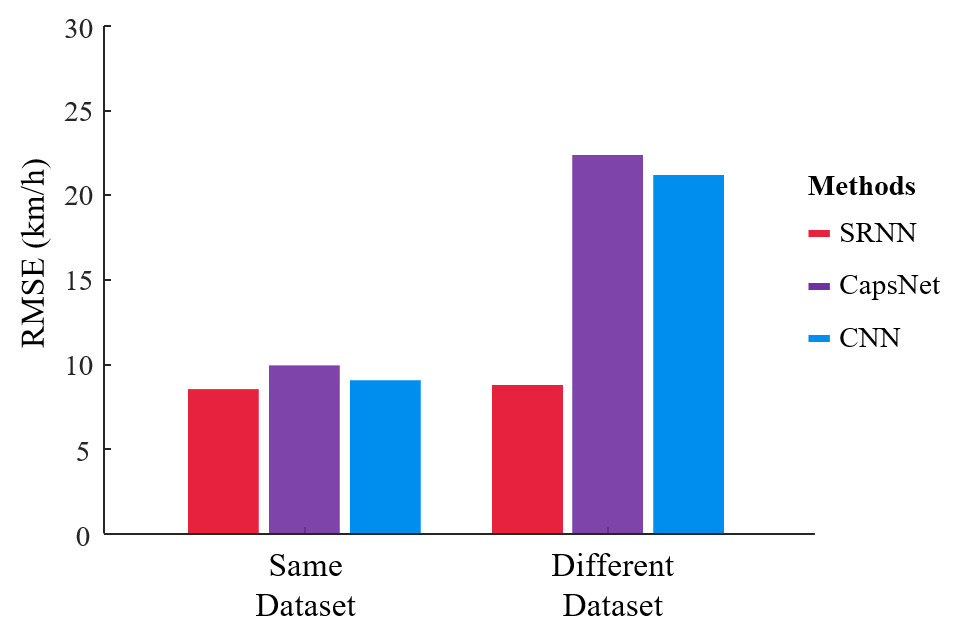}%
	\caption{Average performance of methods in RMSE, evaluated with the same dataset and the different dataset with the training dataset.}
\end{figure}

\begin{figure}[t!]
	\centering
	\includegraphics[clip,width=0.95\linewidth]{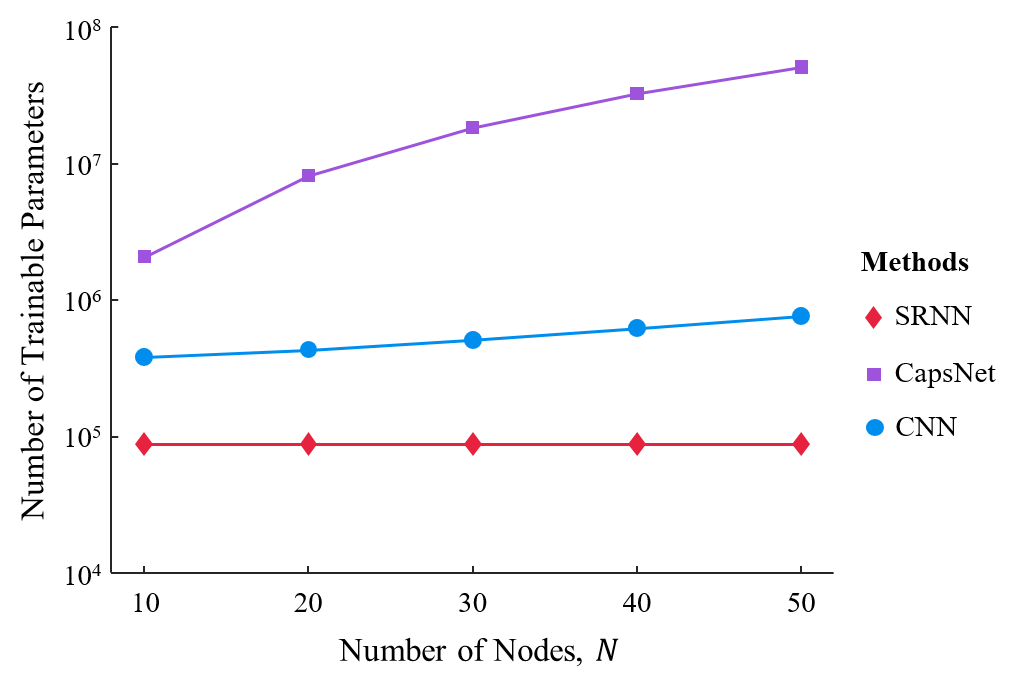}%
	\caption{Number of trainable parameters of the SRNN, CapsNet, and CNN.}
\end{figure}

On the other hand, the image-based methods are not able to predict traffic data for datasets having the different number of road segments. Moreover, even if the number of road segments is equal for a training set and an evaluation set extracted from different datasets, it is obvious that the image-based methods will fail to provide accurate predictions because the image-based methods learn the specific features of the spatio-temporal images that are unique for each dataset. Table \ref{table:scalability} shows that the predictions of CapsNet and CNN on different datasets are erroneous. Fig. 5 shows the average RMSEs of the diagonal elements and off-diagonal elements for the three methods in Table \ref{table:scalability}. The SRNN shows similar performance on the same dataset and the different datasets whereas the CapsNet and CNN show a large difference in the performance between the two cases.

We compare the number of trainable parameters in Fig. 6. As the number of nodes, $N$, increases, the number of trainable parameters of the SRNN remains the same while those of the CapsNet and the CNN keep increasing. When $N=50$, the number of trainable parameters of the CapsNet is over $100$ times larger than that of the SRNN. 

Therefore, the SRNN shows better scalability than the image-based methods in terms of the number of road segments and the applicability to other road networks that have the different number of road segments and different topologies.

\subsection{Discussion}
As one can observe in Section \ref{sec:learning_st} and Section \ref{sec:scalability}, the proposed SRNN outperforms the image-based methods and is applicable to different datasets with different network topologies. This is contributed by the network design where the dynamics of the spatial and temporal edge features learned by the SRNN can be generalized. The result implies that the SRNN is scalable between disjoint road networks (e.g., R1 and R2 where R1$\:\cap\: $R2$\:=\emptyset$) and also between a subset and a superset (e.g., R1 and R4 where R1$\:\subset\:$R4). Besides, the evaluations on road network R4 verify that the SRNN is able to predict traffic data in a road network that is not a connected set. 

Moreover, the number of trainable parameters of the SRNN is independent of the number of the road segments. The size of each RNN, not the number of nodes, determines the number of trainable parameters of the SRNN. The SRNN is able to provide accurate predictions with the smaller number of parameters to train, which will become more advantageous for larger networks.

\section{Conclusion}

In this paper We present an SRNN architecture that learns the spatial and temporal relationship between the traffic data represented as a spatio-temporal graph. To broaden the applicability, we assume that we are using only historical data and the topological information expressed as an adjacency matrix. With the real traffic speed data measured by induction loop detectors in the city of Santander, we demonstrate that the proposed SRNN is computationally more efficient than image-based state-of-the-art approaches. More importantly, we show that the proposed approach is scalable. When the proposed SRNN model is trained with historic road traffic data, the proposed SRNN model shows reliable and accurate performance also over different road networks. The proposed model can be integrated into the traffic management systems or route planning systems to provide accurate predictions on future traffic states by an efficient and scalable manner.

% or


\section*{Acknowledgment}

The authors appreciate the support of the SETA project funded by the European Union’s Horizon 2020 research and innovation program under grant agreement no. 688082.

\ifCLASSOPTIONcaptionsoff
  \newpage
\fi



%\bibliographystyle{IEEEtran}
%\begin{thebibliography}{1}
%\end{thebibliography}

\bibliographystyle{IEEEtran}
\bibliography{IEEESensors}

\begin{thebibliography}{10}

\bibitem{Lv2015}
Yisheng Lv, Yanjie Duan, Wenwen Kang, Zhengxi Li, and Fei-Yue Wang,
\newblock ``Traffic flow prediction with big data: A deep learning approach,''
\newblock {\em IEEE Transactions on Intelligent Transportation Systems}, vol.
  16, no. 2, pp. 865--873, 2015.

\bibitem{Zhang2017}
Junbo Zhang, Yu~Zheng, and Dekang Qi,
\newblock ``Deep spatio-temporal residual networks for citywide crowd flows
  prediction,''
\newblock in {\em Proceedings of the AAAI Conference on Aritificial
  Intelligence}, 2017, pp. 1655--1661.

\bibitem{Ma2015}
Xiaolei Ma, Haiyang Yu, Yunpeng Wang, and Yinhai Wang,
\newblock ``Large-scale transportation network congestion evolution prediction
  using deep learning theory,''
\newblock {\em PloS one}, vol. 10, no. 3, pp. 1--17, 2015.

\bibitem{Wu2018}
Yuankai Wu, Huachun Tan, Lingqiao Qin, Bin Ran, and Zhuxi Jiang,
\newblock ``A hybrid deep learning based traffic flow prediction method and its
  understanding,''
\newblock {\em Transportation Research Part C: Emerging Technologies}, vol. 90,
  pp. 166--180, 2018.

\bibitem{Ma2017}
Xiaolei Ma, Zhuang Dai, Zhengbing He, Jihui Ma, Yong Wang, and Yunpeng Wang,
\newblock ``Learning traffic as images: a deep convolutional neural network for
  large-scale transportation network speed prediction,''
\newblock {\em Sensors}, vol. 17, no. 4, pp. 818, 2017.

\bibitem{Kim2018}
Youngjoo Kim, Peng Wang, Yifei Zhu, and Lyudmila Mihaylova,
\newblock ``A capsule network for traffic speed prediction in complex road
  networks,''
\newblock in {\em Proceedings of the Symposium Sensor Data Fusion: Trends,
  Solutions, and Applications}, 2018.

\bibitem{Xie2010}
Yuanchang Xie, Kaiguang Zhao, Ying Sun, and Dawei Chen,
\newblock ``Gaussian processes for short-term traffic volume forecasting,''
\newblock {\em Transportation Research Record}, vol. 2165, no. 1, pp. 69--78,
  2010.

\bibitem{Chen2015}
Jie Chen, Kian~Hsiang Low, Yujian Yao, and Patrick Jaillet,
\newblock ``Gaussian process decentralized data fusion and active sensing for
  spatiotemporal traffic modeling and prediction in mobility-on-demand
  systems,''
\newblock {\em IEEE Transactions on Automation Science and Engineering}, vol.
  12, no. 3, pp. 901--921, 2015.

\bibitem{Wang2018}
Peng Wang, Youngjoo Kim, Lubos Vaci, Haoze Yang, and Lyudmila Mihaylova,
\newblock ``Short-term traffic prediction with vicinity gaussian process in the
  presence of missing data,''
\newblock in {\em Proceedings of the Symposium Sensor Data Fusion: Trends,
  Solutions, and Applications}, 2018.

\bibitem{Jain2016}
Ashesh Jain, Amir~R Zamir, Silvio Savarese, and Ashutosh Saxena,
\newblock ``Structural-rnn: Deep learning on spatio-temporal graphs,''
\newblock in {\em Proceedings of the Conference on Computer Vision and Pattern
  Recognition}, 2016, pp. 5308--5317.

\bibitem{Vemula2018}
Anirudh Vemula, Katharina Muelling, and Jean Oh,
\newblock ``Social attention: Modeling attention in human crowds,''
\newblock in {\em Proceedings of the International Conference on Robotics and
  Automation}, 2018, pp. 1--7.

\bibitem{ICCASP}
Youngjoo Kim, Peng Wang, and Lyudmila Mihaylova,
\newblock ``Structural recurrent neural network for traffic speed prediction,''
\newblock in {\em Proceedings of the International Conference on Acoustics,
  Speech, and Signal Processing}, 2019.

\bibitem{SETA2016-2}
SETA EU Project, A ubiquitous data and service ecosystem for better
  metropolitan mobility, Horizon 2020 Programme, 2016. Available:
  http://setamobility.weebly.com/.

\bibitem{Zheng2016}
X~Zheng, W~Chen, P~Wang, D~Shen, S~Chen, X~Wang, Q~Zhang, and L~Yang,
\newblock ``Big data for social transportation,''
\newblock {\em IEEE Transactions on Intelligent Transportation Systems}, vol.
  17, no. 3, pp. 620--630, 2016.

\bibitem{Brendel2011}
W~Brendel and S~Todorovic,
\newblock ``Learning spatiotemporal graphs of human activities,''
\newblock in {\em Proceedings of International Conference on Computer Vision
  (ICCV)}, 2011.

\bibitem{Kschischang2001}
Frank~R Kschischang, Brendan~J Frey, and H-A Loeliger,
\newblock ``Factor graphs and the sum-product algorithm,''
\newblock {\em IEEE Transactions on information theory}, vol. 47, no. 2, pp.
  498--519, 2001.

\bibitem{Kingma2014}
Diederik~P Kingma and Jimmy Ba,
\newblock ``Adam: A method for stochastic optimization,''
\newblock {\em arXiv preprint arXiv:1412.6980}, 2014.

\bibitem{Sak2014}
H~Sak, A~Senior, and F~Beaufays,
\newblock ``Long short-term memory based recurrent neural network architectures
  for large vocabulary speech recognition,''
\newblock {\em arXiv preprint arXiv:1402.1128.}, 2014.

\bibitem{He2015}
K~He and J~Sun,
\newblock ``Convolutional neural networks at constrained time cost,''
\newblock in {\em Proceedings of the IEEE conference on computer vision and
  pattern recognition}, 2015, pp. 5353--5360.

\bibitem{Sabour2017}
S~Sabour, N~Frosst, and G.~E. Hinton,
\newblock ``Dynamic routing between capsules,''
\newblock in {\em Advances in Neural Information Processing Systems}, 2017.

\end{thebibliography}





\end{document}